%% file: main_dc.tex
\documentclass[a4paper,fleqn]{cas-dc}

\usepackage[authoryear,longnamesfirst]{natbib}

\def\tsc#1{\csdef{#1}{\textsc{\lowercase{#1}}\xspace}}
\tsc{WGM}
\tsc{QE}

\usepackage{amsmath}
\usepackage{amsfonts}				
\usepackage{mathrsfs} 				

\usepackage{bm}					
\usepackage{textcomp}				

\usepackage[hang,small,bf]{caption}
\usepackage[subrefformat=parens]{subcaption}
\usepackage{wrapfig}

\usepackage{color}

\usepackage{caption}
\usepackage{subcaption}

\usepackage{adjustbox}	

\usepackage{enumitem}		

\usepackage{float}

\begin{document}
\let\WriteBookmarks\relax
\def\floatpagepagefraction{1}
\def\textpagefraction{.001}

\shorttitle{\input{01a_short_title}}    

\shortauthors{\input{01b_short_authors}}  

\title [mode = title]{\input{01_title}}  




\input{00_authors}

\input{00_abstract}




\input{00_keywords}

\maketitle


\section{Introduction}

\input{02_introduction}

\section{Materials}

\input{03_materials}

\section{Methods}

\input{04_methods}

\section{Results}

\input{05_results}

\section{Discussion}

\input{06_discussion}

\section{Conclusion}

\input{07_conclusion}

\input{09_acknowlegements}

\input{10_declaration_of_generative_AI}

\printcredits

\bibliographystyle{cas-model2-names}

\bibliography{ref}



\newpage

\section*{Appendix}
\input{20_appendix}

\end{document}

%% file: 01a_short_title.tex
Cell-Level Histopathological Analysis under Small-Patch Constraints

%% file: 01b_short_authors.tex
H. Kagiyama and T. Nagasaka

%% file: 01_title.tex

Revisiting the Role of Foundation Models in Cell-Level Histopathological Image Analysis under Small-Patch Constraints
 --- Effects of Training Data Scale and Blur Perturbations on CNNs and Vision Transformers

%% file: 00_authors.tex
%

\author[1]{Hiroki Kagiyama}



\ead{kagiyama@med.kobe-u.ac.jp}


\credit{Investigation, Formal analysis, Writing – original draft}

\affiliation[1]{organization={Division of Gastrointestinal Surgery, Department of Surgery,
Kobe University Graduate School of Medicine},
            addressline={7-5-2 Kusunoki-cho, Chuo-ku,}, 
            city={Kobe},
            citysep={}, 
            postcode={650-0017}, 
            state={Hyogo},
            country={Japan}}

\author[2,1]{Toru Nagasaka}[orcid=0009-0002-4322-4023]

\cormark[1]


\ead{toru-ngy@umin.ac.jp}


\credit{Conceptualization, Methodology, Software, Data curation, Visualization, Validation, Writing – review and editing}

\affiliation[2]{organization={Association of Medical Artificial Intelligence Curation},
            addressline={505, Sakae Members Office Building, 4-16-8 Sakae, Naka-ku}, 
            city={Nagoya},
            citysep={}, 
            postcode={460-0008}, 
            state={Aichi},
            country={Japan}}

\author[1]{Yukari Adachi}
\credit{Investigation, Data curation}

\author[1]{Takaaki Tachibana}
\credit{Investigation, Data curation}

\author[1]{Ryota Ito}
\credit{Investigation, Data curation}

\author[3]{Mitsugu Fujita}


\credit{Writing – review and editing}

\affiliation[3]{organization={Center for Medical Education and Clinical Training, Kindai University Faculty of Medicine},
            addressline={1-14-1 Miharadai, Minami-ku}, 
            city={Sakai},
            citysep={}, 
            postcode={590-0111}, 
            state={Osaka},
            country={Japan}}

\author[4,1]{Kimihiro Yamashita}

\affiliation[4]{organization={Department of Biophysics, Kobe University Graduate School of Health Sciences},
            addressline={7-10-2 Tomogaoka, Suma-ku}, 
            city={Kobe},
            citysep={}, 
            postcode={654-0142}, 
            state={Hyogo},
            country={Japan}}

\credit{Resources, Project administration}

\author[1]{Yoshihiro Kakeji}

\credit{Supervision}

\cortext[1]{Corresponding author}

\fntext[1]{}


%% file: 00_abstract.tex
\begin{abstract}
\relax
\textbf{Background and objective:}
Cell-level pathological image analysis requires working with extremely small image patches (40$\times$40 pixels), far below standard ImageNet resolutions.
It remains unclear whether modern deep learning architectures and foundation models can learn robust and scalable representations under this constraint.
We systematically evaluated architectural suitability and data-scale effects for small-patch cell classification.
\textbf{Methods:}
We analyzed 303 colorectal cancer specimens with CD103/CD8 immunostaining, generating 185,432 annotated cell images.
Eight task-specific architectures were trained from scratch at multiple data scales (FlagLimit: 256--16,384 samples per class), and three foundation models were evaluated via linear probing and fine-tuning after resizing inputs to 224$\times$224 pixels.
Robustness to blur was assessed using pre- and post-resize Gaussian perturbations.
\textbf{Results:}
Task-specific models improved consistently with increasing data scale, whereas foundation models saturated at moderate sample sizes.
A Vision Transformer optimized for small patches (CustomViT) achieved the highest accuracy, outperforming all foundation models with substantially lower inference cost.
Blur robustness was comparable across architectures, with no qualitative advantage observed for foundation models.
\textbf{Conclusion:}
For cell-level classification under extreme spatial constraints, task-specific architectures are more effective and efficient than foundation models once sufficient training data are available.
Higher clean accuracy does not imply superior robustness, and large pre-trained models offer limited benefit in the small-patch regime.
\end{abstract}

%% file: 00_keywords.tex
\begin{keywords}
deep learning \sep pathology \sep artificial intelligence \sep cell-level image analysis \sep small-patch \sep foundation model \sep vision transformer \sep CNNs
\end{keywords}

%% file: 02_introduction.tex
In computer vision tasks, convolutional neural networks (CNNs) have achieved remarkable success with standardized input dimensions. Most state-of-the-art architectures—including VGG \citep{simonyan2015deepconvolutionalnetworkslargescale}, ResNet \citep{he2016deep}, and MobileNet \citep{sandler2018mobilenetv2, howard2019searching}—were designed and evaluated using 224×224 pixel inputs for ImageNet classification. Even Inception V3 \citep{szegedy2015rethinkinginceptionarchitecturecomputer}, which uses a slightly larger 299×299 pixel input, maintains this paradigm of relatively large input dimensions. These specific dimensions became the de facto standard for three key reasons: first, these models were trained on ImageNet with these input sizes and their pre-trained weights are publicly available; second, performance benchmarks and comparisons across models are consistently reported at these dimensions; and third, transfer learning applications typically maintain these original input sizes to leverage the learned feature representations effectively.

In digital pathology, the analysis of whole slide images (WSIs) typically employs a patch-based approach due to the gigapixel nature of these images. Standard practice involves extracting patches of 224×224 pixels or larger (commonly 400×400 to 500×500 pixels) from WSIs, which are then processed through CNNs \citep{meirelles2022building, srinidhi2021deep}. These patch sizes, often corresponding to approximately 50×50 $\mu m$ of tissue area, are designed to capture sufficient contextual information for accurate classification.

This study presents a fundamentally different challenge: analyzing patches of merely 40×40 pixels, approximately 1/30 the area of standard ImageNet inputs. This extreme constraint emerges from the practical requirement of single-cell level analysis in pathological diagnosis, where each patch may contain only one or a few cells. Unlike conventional approaches that leverage rich contextual information from larger tissue regions, our method must extract diagnostic features from minimal spatial information.

\begin{figure}[htbp]
\centering
\includegraphics[width=\columnwidth]{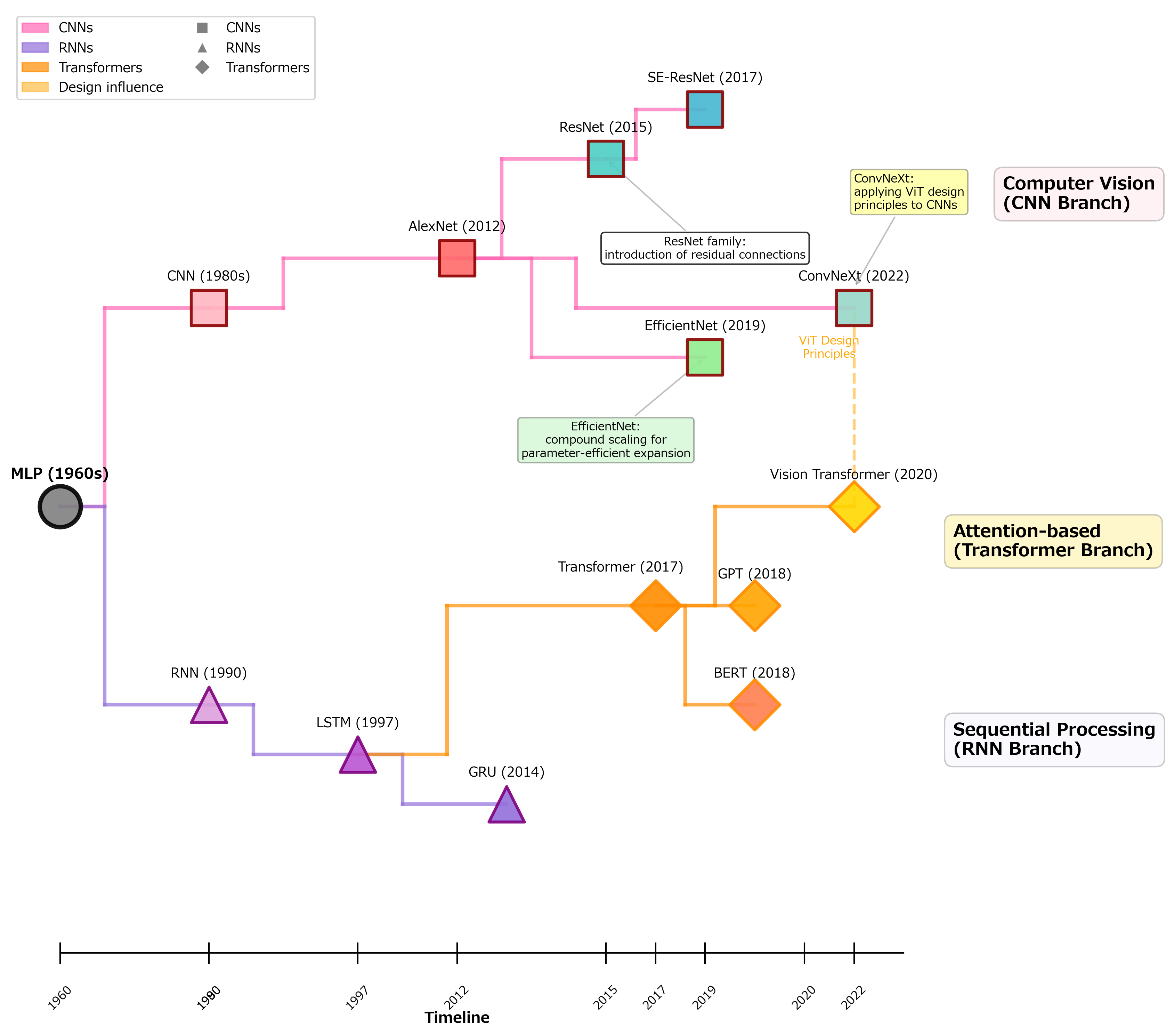}
\caption{
A schematic phylogenetic overview of major deep learning architectures.
Early multilayer perceptrons gave rise to convolutional neural networks, which evolved through deeper residual designs and efficiency-oriented variants.
In parallel, recurrent models led to attention-based transformers, from which Vision Transformer emerged as a direct adaptation to visual tasks.
The diagram also illustrates how transformer design principles subsequently influenced modern convolutional architectures such as ConvNeXt, reflecting cross-paradigm convergence in recent model development.
}
\label{fig:phylogenetic_tree}
\end{figure}

To systematically investigate this challenge, we implemented a comprehensive set of architectures spanning multiple design paradigms.
These architectures are situated within the broader historical context of deep learning model evolution, as summarized in Figure~\ref{fig:phylogenetic_tree}.
These include baseline models (MLP, CNN, ResNet) and modern architectures representing recent advances: Squeeze-and-Excitation Networks \citep{hu2019squeezeandexcitationnetworks}, which enhance channel interdependencies through adaptive recalibration; EfficientNet \citep{tan2020efficientnetrethinkingmodelscaling}, which employs compound scaling; Vision Transformer \citep{dosovitskiy2021imageworth16x16words}, which applies self-attention mechanisms; and ConvNeXt \citep{liu2022convnet2020s}, which modernizes ConvNets with transformer-inspired designs.
Additionally, we evaluated foundation models through linear probing and fine-tuning to assess their transferability to small-patch learning.

To our knowledge, no systematic study has compared these modern architectures under such extreme spatial constraints in pathological image analysis. While existing literature extensively covers various patch sizes for WSI analysis, the lower bound of viable patch dimensions and the architectural characteristics that enable learning from minimal spatial information remain unexplored. This gap is particularly significant given that architectural designs and learned features optimized for 224×224 inputs may not translate effectively to such small scales.

Meanwhile, the current trend in computer vision and medical imaging is to leverage large-scale foundation models through fine-tuning or linear probing. This paradigm, however, presupposes that the input scale and visual features of the downstream task are compatible with those of the pretraining corpus. In our case, the 40×40 pixel patches used for cell-level pathology analysis lie far outside the scale regime for which ImageNet-pretrained models were optimized. Simply resizing such inputs to 224×224 either destroys fine morphological details or introduces artifacts that are irrelevant to the pretrained feature space. Consequently, the transferability of foundation model representations is severely limited, and training directly on domain-specific, low-resolution cell patches becomes indispensable. 

This work addresses three critical questions: First, can modern deep learning architectures learn meaningful representations from 40×40 pixel pathological images? Second, which architectural paradigms—convolutional, attention-based, or hybrid—are most suitable for such small inputs? Third, how do practical constraints such as blur and focus issues affect model performance at this scale? 
Through systematic experimentation with these architectures, we demonstrate that despite the severe spatial constraints, certain architectural designs can achieve clinically relevant accuracy in cell classification tasks, with significant differences in their effectiveness at this scale.
Furthermore, our results provide an empirical perspective on the boundaries of foundation-model-based transfer learning in histopathological image analysis.

%% file: 03_materials.tex
\subsection{Patients and Samples}

This study included tissue specimens from 303 colorectal cancer patients who underwent biopsy or surgery, with or without neoadjuvant chemoradiotherapy. Both biopsy and surgical samples were formalin-fixed and paraffin-embedded, and were subjected to dual immunostaining for CD103 and CD8. The staining process was performed according to established protocols to ensure consistency and reliability \citep{tachibana2025reliability, ohno2024tumor}.

\subsection{Data Annotation Methodology Using Cu-Cyto\textsuperscript{\tiny{\textregistered}} Viewer}

Cell-level annotation was performed using the Cu-Cyto\textsuperscript{\tiny{\textregistered}} Viewer, a specialized annotation tool designed for precise cellular identification in histopathological images \citep{abe2023deep}. The annotation workflow employed a human-AI collaborative approach to ensure comprehensive cellular coverage and annotation accuracy across all tissue sections.

The annotation protocol was essentially identical to that described in \citep{tachibana2025reliability}; therefore, only a brief summary is provided here. Initially, a prototype AI model performed automated cell detection and placed preliminary annotation markers across the tissue sections. Subsequently, expert annotators systematically reviewed the AI-generated annotations using the Cu-Cyto\textsuperscript{\tiny{\textregistered}} Viewer interface, verifying markers for accuracy and adding markers for cells missed by the automated detection system.

Annotation quality was maintained through a multi-stage verification protocol. Following the initial annotation refinement by expert annotators, a second expert annotator performed a double-check of the annotations, with a final review conducted by a pathologist to ensure the highest level of accuracy and reliability. Inter-annotator agreement was assessed through systematic comparison of overlapping annotation regions, ensuring consistent application of classification criteria across all tissue sections.

The Cu-Cyto\textsuperscript{\tiny{\textregistered}} Viewer platform recorded annotation metadata including marker coordinates and annotator identification for traceability. This protocol required comprehensive cellular coverage, ensuring that every identifiable cell within the tissue section received appropriate classification markers, which serves as the basis for estimating true prior probabilities for cell classification.

\subsection{Dataset Preparation and Sampling Strategy}

To address class imbalance inherent in the original dataset and create balanced training sets suitable for deep learning, we implemented a systematic data selection and augmentation pipeline. The process consisted of class-balanced sampling and data augmentation through geometric transformations.

\subsubsection{Class-Balanced Sampling}

The class-balanced sampling procedure used in this study was essentially identical to that described in \citep{tachibana2025reliability}; therefore, only a brief summary is provided here.

Data selection was performed using a custom Python implementation that applies stratified sampling to achieve uniform class distribution. The sampling strategy was controlled by two key parameters: \texttt{FlagLimit}, which defines the maximum number of samples per class, and \texttt{FlagMin}, set to 5\% of \texttt{FlagLimit}, which defines the minimum threshold for class inclusion.

\subsubsection{Sampling Analysis and Validation}

The effectiveness of the sampling strategy was evaluated through systematic analysis of the selection process across all seven \texttt{FlagLimit} conditions (256, 512, 1024, 2048, 4096, 8192, and 16384). From an original dataset of 185,432 samples across 20 unique flags, the sampling process achieved substantial data reduction while maintaining class balance.

%% file: 04_methods.tex
\subsection{Dataset and Experimental Setup}

Dataset details including patient cohorts, cell type classifications, and class-balanced sampling strategy are described in the Materials section.
For all experiments, we used the balanced datasets generated at seven \texttt{FlagLimit} values (256, 512, 1024, 2048, 4096, 8192, and 16384).
Each \texttt{FlagLimit} value determines the maximum number of samples per class used for training, resulting in balanced datasets with varying total sizes depending on the number of classes meeting the inclusion threshold.
Each dataset was randomly split into training (80\%) and validation (20\%) subsets using stratified sampling to maintain class balance.
All models were evaluated on the same fixed validation set for each \texttt{FlagLimit} condition.

\subsection{Data Augmentation}
\label{sec:method_data_aug}

A common set of geometric transformations was applied exhaustively to every
training sample: horizontal flip, vertical flip, and diagonal flip (transpose).
These three transformations, combined with the original orientation, produce
four copies of each sample, yielding a $4\times$ expansion of the training set
prior to any color- or pixel-domain augmentation.

In addition, five color-space transformations were introduced:
(i)~R-channel $\gamma$ correction ($\gamma = 1.2$), applied to 20\% of the
original training samples;
(ii)~B-channel $\gamma$ correction ($\gamma = 1.2$), applied to 30\%;
(iii)~full-channel (RGB) $\gamma$ correction ($\gamma = 1.2$), applied to 50\%;
(iv)~HSV saturation--value transformation ($\alpha_S = 0.8$, $\alpha_V = 1.1$),
applied to 20\%;
and (v)~a combined HSV~+~B-channel transformation
($\alpha_S = 0.7$, $\alpha_V = 1.1$, B\,$\gamma = 1.3$), applied to 20\%.
Each transformed subset was derived from the \emph{original} (pre-geometric)
training samples and subsequently subjected to the same fourfold geometric
expansion, so the total dataset size after all augmentations is
$4 \times (1 + 0.20 + 0.30 + 0.50 + 0.20 + 0.20) = 4 \times 2.40 = 9.6\times$
the original sample count.

\subsection{Task-Specific Model Architectures}

To systematically evaluate the performance of models trained from scratch under small-patch constraints, we implemented seven task-specific architectures representing diverse design paradigms. All models were designed or adapted for $40 \times 40$ pixel RGB inputs (3 channels) and trained using Adam optimizer with cross-entropy loss.

\textbf{MLP.}
The Multi-Layer Perceptron consists of fully connected layers with hidden dimensions [4800, 1500, 800, 600] and tanh activation. Input images ($40 \times 40 \times 3 = 4800$ features) are flattened before processing. Input dropout (0.1) and hidden dropout (0.5) were applied during training. Total parameters: 3.37M.

\textbf{CNN.}
The CNN comprises four convolutional blocks: Conv1 ($3 \times 3$, 48 channels), Conv2 ($3 \times 3$, 64 channels), Conv3 ($3 \times 3$, 128 channels), and Conv4 ($3 \times 3$, 128 channels), each followed by ReLU activation, $2\times2$ max pooling (stride 2), and dropout (0 for convolutional layers by default). The spatial dimension reduces from $40 \times 40 \rightarrow 20 \times 20 \rightarrow 10 \times 10 \rightarrow 5 \times 5$. Two fully connected layers (3200 $\rightarrow$ 512 $\rightarrow$ 256) with tanh activation process the flattened features before the classification head. Input dropout: 0.1, hidden dropout: 0.5. Total parameters: 2.02M.

\textbf{ResNet-D4.}
ResNet-D4 is a custom residual network with four blocks of progressively increasing channels (24 $\rightarrow$ 48 $\rightarrow$ 64 $\rightarrow$ 128). Each block contains 1 residual unit by default (configurable based on total block count). Each residual unit consists of two $3 \times 3$ convolutional layers with ReLU activation and skip connections. After the first block, original RGB channels are concatenated with feature maps before transition to the second block. Fully connected layers (3200 $\rightarrow$ 512 $\rightarrow$ 256) with tanh activation precede the classification head. Input dropout: 0.1, hidden dropout: 0.5, convolutional dropout: 0. Total parameters: 2.27M.

\textbf{NIN (Network-in-Network).}
NIN is a dual-pathway hybrid architecture combining ResNet-D4 and MLP branches. The ResNet pathway uses the same structure as ResNet-D4 but with separate fully connected layers (3200 $\rightarrow$ 512 $\rightarrow$ 256). The MLP pathway consists of layers with dimensions [4800, 1500, 800, 600], initialized with autoencoder-learned weights when available. Outputs from both pathways (256 + 600 = 856 features) are concatenated before the final classification layer. This design aims to leverage both spatial hierarchical features and global statistical patterns. Total parameters: 7.41M.

\textbf{SE-ResNet-D4.}
SE-ResNet-D4 extends ResNet-D4 by inserting Squeeze-and-Excitation (SE) blocks at five strategic locations: after transitions to Block 3 (64 channels) and Block 4 (128 channels), at the end of Block 3 and Block 4, and after the final convolution. Each SE block performs channel-wise recalibration through global average pooling, two fully connected layers with reduction ratio 2 (e.g., 128 $\rightarrow$ 64 $\rightarrow$ 128 for 128 channels), ReLU activation, SE dropout (0.1), and sigmoid gating. The base ResNet-D4 structure remains unchanged. Total parameters: 2.33M.

\textbf{EfficientNet-B0.}
EfficientNet-B0 is adapted for $40\times40$ inputs. The stem uses $3\times3$ convolution (stride 1, padding 1) projecting to 32 channels. Seven MBConv stages with configuration [expansion ratio, output channels, number of blocks, stride]: [1, 16, 1, 1], [6, 24, 2, 2], [6, 40, 2, 2], [6, 80, 3, 2], [6, 112, 3, 1], [6, 192, 4, 2], [6, 320, 1, 1]. Each MBConv block employs expansion via $1\times1$ convolution, $3\times3$ depthwise separable convolution, SE block (reduction ratio 0.25, SE dropout 0.1), and $1\times1$ projection. The head uses $1\times1$ convolution to 1280 channels, global average pooling, and linear classifier. Batch normalization and SiLU activation are used throughout. Total parameters: 3.93M. Learning rate: $1 \times 10^{-4}$.

\textbf{ConvNeXt-Tiny.}
ConvNeXt-Tiny is adapted with reduced dimensions and kernel sizes. The stem uses $2\times2$ convolution (stride 2) projecting to 48 channels. Four stages with depths [2, 2, 6, 2] and channel dimensions [48, 96, 192, 384] employ ConvNeXt blocks. Each block contains: LayerNorm, $5\times5$ depthwise convolution (reduced from standard $7\times7$), LayerNorm, $1\times1$ expansion to $4 \times$ channels, GELU activation, $1\times1$ projection back, and residual connection. Downsampling between stages uses LayerNorm followed by $2\times2$ strided convolution. Drop path rate: 0.2. Global average pooling precedes the classification head. Total parameters: 4.78M. Learning rate: $1 \times 10^{-4}$, weight decay: 0.2.

\textbf{CustomViT.}
CustomViT divides $40\times40$ images into $8\times8$ pixel patches, yielding $5 \times 5 = 25$ patch tokens. Each patch is linearly projected to 160-dimensional embeddings. A learnable class token and positional embeddings (160-dimensional) are added to the sequence. Six transformer blocks process the token sequence, each containing: LayerNorm, multi-head self-attention (4 heads, 40 dimensions per head), residual connection, LayerNorm, MLP with hidden dimension $160 \times 4 = 640$, GELU activation, and residual connection. Dropout rate: 0.2 (applied after attention and MLP). The final class token output after LayerNorm feeds into the classification head. Weights initialized with truncated normal distribution (std=0.02). Total parameters: 1.89M. Learning rate: $1 \times 10^{-4}$, weight decay: 0.01.

All models were trained with batch size 512, using early stopping based on validation accuracy with patience of 10 epochs. Training epochs varied by architecture: 50 epochs for ResNet-based models (ResNet-D4, NIN, SE-ResNet-D4), 100 epochs for CNN, 30 epochs for MLP, 80 epochs for CustomViT, 60 epochs for ConvNeXt-Tiny, and 50 epochs for EfficientNet-B0. Input dropout (0.1) and hidden dropout (0.5) were applied to MLP, CNN, ResNet-D4, NIN, and SE-ResNet-D4 during training.

\subsection{Foundation Models: Linear Probing and Fine-Tuning}

\textbf{Input Adaptation.}
To apply foundation models pretrained on $224 \times 224$ images to our $40 \times 40$ cell patches, 
input images were resized to $224 \times 224$ using bicubic interpolation, maintaining the aspect ratio.

\textbf{Linear Probing (LP).}
In linear probing, all pretrained backbone weights were frozen, and only the final classification head 
(a single linear layer projecting from the model's feature dimension to the number of classes) was trained. 
The classification head was randomly initialized and trained for 50 epochs with learning rate $5 \times 10^{-4}$, 
batch size 256, and Adam optimizer.

\textbf{Fine-Tuning Last Layer (FT\_last).}
For fine-tuning, we unfroze the final transformer block (for ViT-based models) or 
the final stage (for CNN-based models) along with the classification head, 
while keeping all earlier layers frozen. 
Training was performed for 50 epochs with learning rate $1 \times 10^{-4}$, 
batch size 256, weight decay 0.01, and Adam optimizer.

Table~\ref{tab:foundation_model_comparison} summarizes the foundation models examined
in this study. These models, ranging from convolutional to transformer-based architectures,
were evaluated for cell-level classification via linear probing and fine-tuning under
low-resolution constraints.

\begin{table*}[htbp]
\centering
\caption{Foundation models used for cell-level linear probing or fine-tuning}
\label{tab:foundation_model_comparison}
\adjustbox{width=\textwidth}{
\begin{tabular}{lcccccc}
\hline
\textbf{Model} & \textbf{Architecture Type} & \textbf{Input Size} & \textbf{Pretraining Dataset} & \textbf{Year} & \textbf{Parameter Count (M)} & \textbf{Reference / Source} \\
\hline
ResNet-RS50 \citep{bello2021revisiting} & Convolutional Neural Network (CNN) & $224\times224$ & ImageNet-1k & 2021 & 33.7 & \texttt{timm/resnetrs50.tf\_in1k} \\
CTransPath \citep{wang2022transformer} & Swin Transformer & $224\times224$ & 15M histopathology tiles & 2022 & 27.5 & \texttt{1aurent/swin\_tiny\_patch4\_window7\_224.CTransPath} \\
UNI \citep{chen2024uni} & Vision Transformer (Hybrid) & variable ($224$--$448$) & Multi-domain (Histo + Natural) & 2023 & 303.4 & \texttt{hf-hub:MahmoodLab/UNI} \\
\hline
\end{tabular}
}
\end{table*}

\subsection{Optimal Learning Strategies by Dataset Scale.}

\renewcommand{\arraystretch}{2.5}
\begin{table*}[htbp]
\centering
\caption{Empirically suggested optimal learning strategies by dataset scale in cell-level training}
\label{tab:optimal_strategy_by_scale}
\adjustbox{width=\textwidth}{
\begin{tabular}{lccc}
\hline
\makecell[l]{\textbf{Dataset Scale} \\ \textbf{(Number of Patches)}} &
\textbf{Representative Study} &
\textbf{Primary Observation} &
\textbf{Suggested Optimal Strategy} \\
\hline
$<$ 1k patches &
\makecell[l]{\cite{koohbanani2020selfpathselfsupervisionclassificationpathology}\\(Self-Path)} &
\makecell[l]{Self-supervised pretraining provides stable feature representations\\under limited annotations, while direct fine-tuning easily overfits.\\Although linear probing was not explicitly evaluated, it can be\\considered conceptually analogous to this feature-based approach.} &
\makecell[l]{\textbf{Feature Extraction or Linear Probing}\\(frozen pretrained backbone)} \\

1k–10k patches &
\makecell[l]{\cite{rohit2023domain}\\(CLAM / TransMIL)} &
\makecell[l]{Demonstrates that partial adaptation of pre-trained models,\\such as domain-specific fine-tuning of feature extractors,\\substantially improves performance even with moderate data sizes.} &
\makecell[l]{\textbf{Partial Fine-Tuning}\\(last-layer FT)} \\

$>$ 10k patches &
\makecell[l]{\cite{chen2024uni}\\(UNI, MahmoodLab)} &
\makecell[l]{With large-scale datasets, full fine-tuning achieves the best\\generalization, and from-scratch training can also match or surpass it.} &
\makecell[l]{\textbf{Full Fine-Tuning}\\or \textbf{From-Scratch Training}} \\
\hline
\end{tabular}
}
\end{table*}
\renewcommand{\arraystretch}{1.0} 

The optimal training strategy for cell-level classification varies considerably depending on the number of available image patches (Table \ref{tab:optimal_strategy_by_scale}). 
When the dataset is extremely limited (fewer than $10^3$ patches), 
\citet{koohbanani2020selfpathselfsupervisionclassificationpathology} demonstrated that 
linear probing (LP) offers greater stability under limited annotations, 
whereas fine-tuning tends to overfit due to insufficient diversity of training examples. 
In the intermediate range of approximately $10^3$–$10^4$ patches, 
\citet{rohit2023domain} reported that partial fine-tuning, particularly of the last block or classifier head (``last-layer FT''), 
substantially improves downstream performance compared with pure LP. 
Finally, for large-scale datasets exceeding $10^4$ patches, 
\citet{chen2024uni} showed that full fine-tuning provides the highest generalization capability, 
and that training from scratch can achieve comparable or even superior performance when sufficient data are available. 
Taken together, these findings suggest that the optimal strategy shifts gradually from LP to full fine-tuning 
as the dataset size increases, with a transition region where partial fine-tuning is most beneficial.

However, the applicability of these trends to small cell-level patches (e.g., $40 \times 40$ pixels) remains largely unexplored. 
Most previous studies have focused on tile-level or region-level patches ($\geq224 \times 224$ pixels), 
while the behavior of foundation models under strong spatial downscaling and limited contextual information 
is still uncertain. 
In this study, we aim to systematically investigate the robustness and performance of multiple foundation models 
under such low-resolution, cell-level conditions, comparing linear probing, partial fine-tuning, 
and from-scratch training within a unified evaluation framework.

\subsection{Robustness to Blur}

To assess robustness against image blur under low-resolution constraints,
we evaluated two complementary corruption schemes that differ in the stage
at which blur is introduced:
(i) \emph{pre-resize blur}, applied to the original-resolution image prior to
downsampling, approximating optical defocus or limited depth-of-field effects
during image acquisition; and
(ii) \emph{post-resize blur}, applied after resizing the image to
$40\times40$ pixels, corresponding to digitally introduced degradation at the
model input level.
This distinction allows us to contrast blur effects that are physically
plausible at the acquisition stage with synthetic perturbations commonly used
in robustness analyses.

For both schemes, blur was implemented using an isotropic Gaussian filter.
In the post-resize setting, Gaussian blur with standard deviation $\sigma$ was
applied directly to the resized $40\times40$ RGB image.
In the pre-resize setting, to ensure comparability between the two schemes,
the blur strength was adjusted to account for the difference in spatial
resolution between the original image and the resized patch.
Specifically, if the resized image has spatial size $S$ (here $S=40$) and the
original image has size $S_0$, the effective pre-resize blur standard deviation
was defined as
\[
\sigma_{\mathrm{pre}} = \sigma \times \frac{S_0}{S},
\]
so that the spatial extent of blur is normalized when expressed on the resized
grid.
This correction prevents resolution mismatch from artificially amplifying or
attenuating the apparent effect of pre-resize blur.

After blurring and resizing, all images were converted to channel-first
tensors.
To maintain consistency with the training pipeline, mean subtraction
(whitening) was applied identically to clean and corrupted images.
Whitening was performed only on the RGB channels using the mean image computed
from the training set, while auxiliary component channels, when present, were
left unchanged.
This design isolates the impact of blur on visual information without
introducing confounding normalization effects.

Gaussian blurring was implemented using the
\texttt{ImageFilter.\allowbreak GaussianBlur} function from the Python Imaging Library
(PIL), ensuring deterministic and isotropic filtering.
Blur severity was evaluated over a geometric progression
$\sigma \in \{0.1, 0.2, 0.4, 0.8, 1.6\}$, corresponding to successive doubling of
the blur scale.
Performance degradation was quantified relative to the clean baseline using
accuracy and macro-F1 score.

To summarize robustness trends across this range, we additionally analyzed
the rate of performance degradation as a function of blur magnitude by
estimating the drop slope with respect to $\log_2(1+\sigma)$.
This representation reflects the multiplicative nature of blur severity and
highlights the onset and acceleration of performance degradation observed in
the results.

\subsection{Evaluation Metrics}

Model performance was evaluated using accuracy, macro-averaged precision, recall, and F1 score. 
Macro-averaging computes metrics independently for each class and then averages them, 
treating all classes equally regardless of their support in the dataset. 
This approach is particularly appropriate for balanced datasets and provides a fair assessment 
of model performance across all cell types.

Inference time was measured as the average forward pass time per sample on a single GPU, 
computed over 400 inference runs with batch size 1 after model warm-up.

\subsection{Implementation Details}

All experiments were implemented in Python (v3.8) using NumPy, pandas,
scikit-learn, matplotlib, and PyTorch (v1.7).
Deep learning models were trained using the Adam optimizer with architecture-specific
hyperparameters as described above.
Random seeds were fixed (seed=42) for data splitting to ensure reproducibility.
Computations were performed on a workstation equipped with an NVIDIA RTX~3060
GPU and 64\,GB RAM.

%% file: 05_results.tex
\subsection*{Inference performance}

\begin{figure}[htbp]
\centering
\includegraphics[width=\columnwidth]{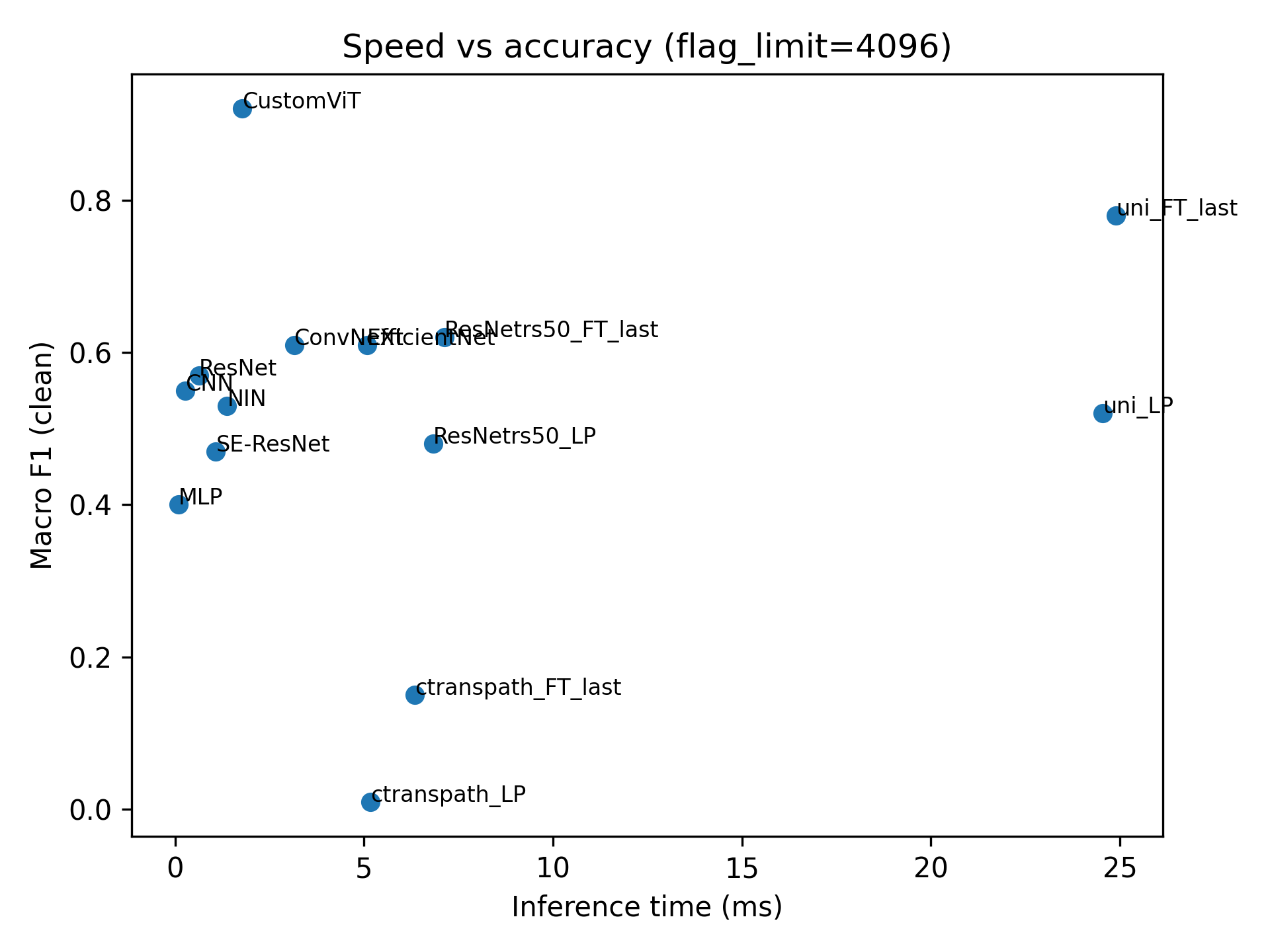}
\caption{
Trade-off between inference speed and classification performance at \texttt{flag\_limit} = 4096.
Each point represents a model's macro-F1 score (y-axis) versus its inference time per patch (x-axis).
CustomViT achieves the highest classification performance while maintaining inference costs more than an order of magnitude lower than foundation models such as UNI.
Task-specific convolutional models (CNN, ResNet, NIN, SE-ResNet) offer sub-millisecond to few-millisecond inference but plateau at substantially lower performance.
Foundation models (ResNet-RS50, CTransPath, UNI) incur markedly higher inference latency, with UNI requiring approximately 25~ms per patch despite strong fine-tuning performance.
This demonstrates that CustomViT provides a favorable balance between accuracy and computational efficiency for large-scale patch-based analysis.
}
\label{fig:speed_vs_accuracy}
\end{figure}

Table~\ref{tab:inference_performance_4096} summarizes the inference-time
characteristics of task-specific and foundation models at a large training
scale (\texttt{flag\_limit} = 4096).
Task-specific models trained from scratch exhibited substantially lower
inference latency and model size compared with foundation models.
Among them, MLP, CNN, and ResNet achieved sub-millisecond inference times,
while more expressive architectures such as CustomViT and ConvNeXt remained
within a few milliseconds.
In contrast, foundation models incurred markedly higher inference costs.
For example, the UNI model required approximately 25~ms per inference and over
1~GB of model storage, reflecting its substantially larger parameter count.
Importantly, CustomViT achieved competitive or superior classification
performance (Fig.~\ref{fig:scaling_flag_limit}) while maintaining an inference
cost more than an order of magnitude lower than that of the largest foundation
models, underscoring its suitability for large-scale patch-based analysis.
The trade-off between inference speed and classification accuracy is visualized
in Figure~\ref{fig:speed_vs_accuracy}, which demonstrates that CustomViT
occupies a uniquely favorable position in the accuracy-efficiency space,
substantially outperforming both compact convolutional models and
computationally expensive foundation models.

\begin{table}[htbp]
\centering
\caption{Inference performance of task-specific and foundation models at large-scale training
(\texttt{flag\_limit} = 4096).}
\label{tab:inference_performance_4096}
\adjustbox{width=\columnwidth}{
\begin{tabular}{lcccc}
\hline
Model & Stage & Inference time (ms) & Parameters (M) & Model size (MB) \\
\hline
MLP & -- & 0.09 & 3.37 & 12.8 \\
CNN & -- & 0.27 & 2.02 & 7.7 \\
ResNet & -- & 0.64 & 2.27 & 8.7 \\
NIN & -- & 1.38 & 7.41 & 28.3 \\
SE-ResNet & -- & 1.08 & 2.33 & 8.9 \\
EfficientNet & -- & 5.08 & 3.93 & 15.1 \\
CustomViT & -- & 1.78 & 1.89 & 7.2 \\
ConvNeXt & -- & 3.15 & 4.78 & 18.3 \\
\hline
ResNet-RS50 & LP & 6.83 & 33.7 & 128.7 \\
ResNet-RS50 & FT\_last & 7.13 & 33.7 & 128.7 \\
CTransPath & LP & 5.17 & 27.5 & 106.1 \\
CTransPath & FT\_last & 6.35 & 27.5 & 106.1 \\
UNI & LP & 24.54 & 303.4 & 1157.3 \\
UNI & FT\_last & 24.89 & 303.4 & 1157.3 \\
\hline
\end{tabular}
}
\end{table}

\subsection*{Classification performance}

\begin{figure}[htbp]
\centering
\includegraphics[width=\columnwidth]{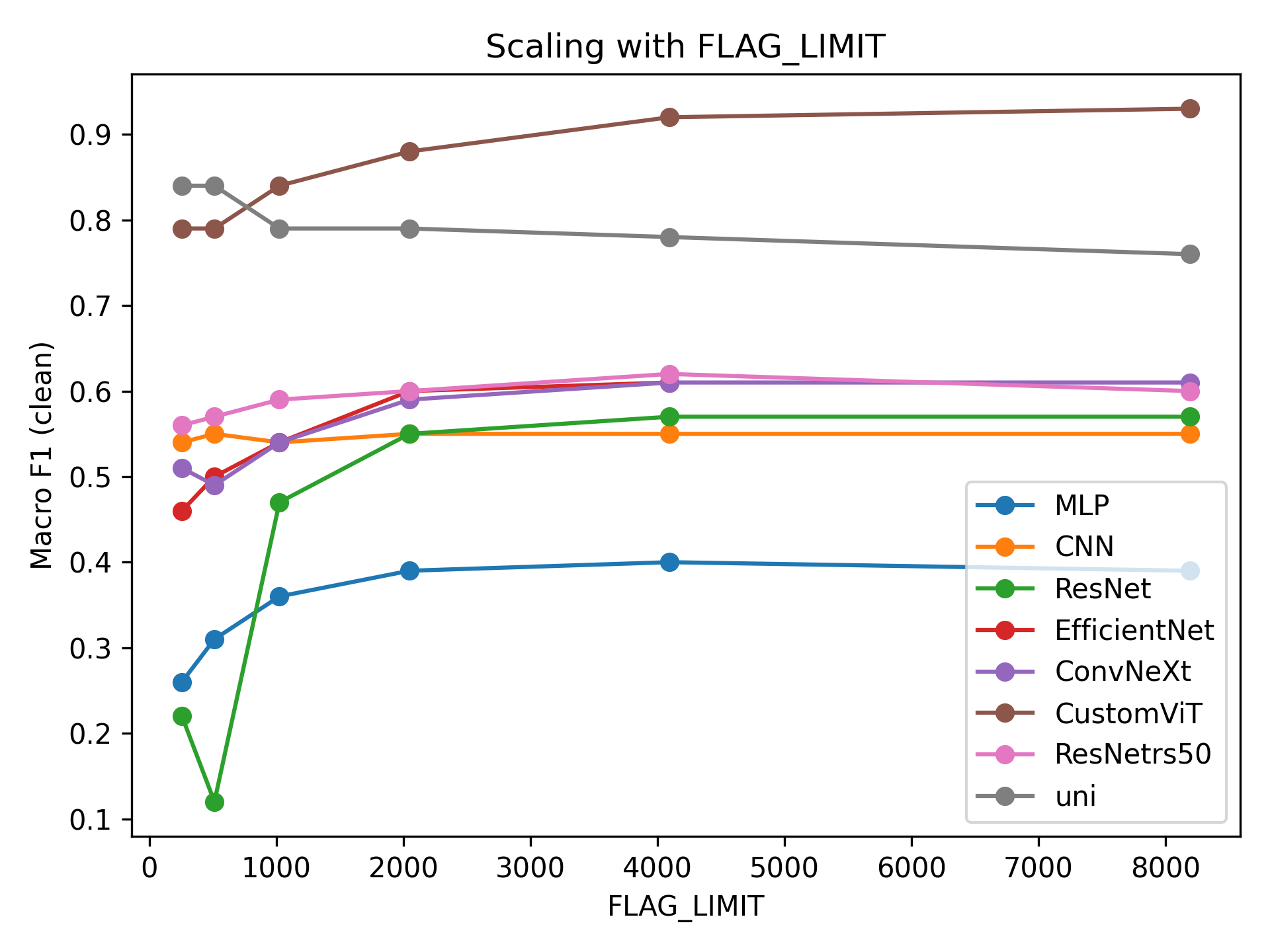}
\caption{
Scaling behavior of classification performance with increasing training set size
(\texttt{flag\_limit}). Macro F1 scores under clean conditions are shown for
task-specific models and foundation models with fine-tuning of the last layer (FT\_last).
CustomViT exhibits monotonic performance gains and surpasses foundation models
at moderate data scales, whereas conventional convolutional networks saturate.
EfficientNet shows early gains but was not evaluated at the largest scale due to
excessive training time.
}
\label{fig:scaling_flag_limit}
\end{figure}

Figure~\ref{fig:scaling_flag_limit} summarizes the scaling behavior of
classification performance as a function of training set size
(\texttt{flag\_limit}).
Foundation models achieved strong performance in the low-data regime, but
showed limited improvement as additional training data were introduced.
In contrast, task-specific models exhibited heterogeneous scaling behaviors.
Among them, CustomViT demonstrated consistent and monotonic performance gains,
surpassing all foundation models at moderate data scales.
Conventional convolutional architectures, including CNN, ResNet, and ConvNeXt,
improved with increasing data but saturated below the performance of the best
foundation models.
EfficientNet showed early performance gains but could not be evaluated at the
largest scale due to prohibitive training cost, highlighting its limited
scalability in large-scale patch-based learning.

Table~\ref{tab:classification_performance_4096} reports the classification
performance of task-specific and foundation models at a representative
large-scale training setting (\texttt{flag\_limit} = 4096).
Among task-specific models trained from scratch, performance increased
consistently with model capacity.
CustomViT achieved the best overall results, attaining an accuracy of 0.92 and
a macro-F1 score of 0.92, clearly outperforming conventional convolutional
architectures.

ConvNeXt, ResNet, and NIN showed moderate improvements with increasing data,
reaching macro-F1 scores in the range of 0.53--0.61.
In contrast, simpler architectures such as MLP and CNN saturated at lower
performance levels.

Among foundation models, fine-tuning (FT\_last) consistently outperformed
linear probing (LP).
UNI with FT\_last achieved a macro-F1 score of 0.78, representing the strongest
foundation-model baseline at this scale.
CTransPath, however, exhibited markedly lower performance (F1 = 0.01 for LP and
0.15 for FT\_last), suggesting limited transferability of this Swin Transformer–based
architecture to the small-patch regime.
Notably, CustomViT already surpassed all foundation models at
\texttt{flag\_limit} = 4096, while maintaining substantially lower inference
costs (Table~\ref{tab:inference_performance_4096}).

\begin{table}[htbp]
\centering
\caption{Classification performance of task-specific and foundation models at
\texttt{flag\_limit} = 4096.}
\label{tab:classification_performance_4096}
\adjustbox{width=\columnwidth}{
\begin{tabular}{lcccc}
\hline
Model & Stage & Accuracy & Macro precision & Macro F1 \\
\hline
MLP & -- & 0.49 & 0.42 & 0.40 \\
CNN & -- & 0.59 & 0.58 & 0.55 \\
NIN & -- & 0.60 & 0.58 & 0.53 \\
ResNet & -- & 0.61 & 0.58 & 0.57 \\
SE-ResNet & -- & 0.53 & 0.50 & 0.47 \\
EfficientNet & -- & 0.64 & 0.62 & 0.61 \\
CustomViT & -- & \textbf{0.92} & \textbf{0.91} & \textbf{0.92} \\
ConvNeXt & -- & 0.65 & 0.63 & 0.61 \\
\hline
ResNet-RS50 & LP & 0.51 & 0.51 & 0.48 \\
ResNet-RS50 & FT\_last & 0.63 & 0.62 & 0.62 \\
CTransPath & LP & 0.09 & 0.01 & 0.01 \\
CTransPath & FT\_last & 0.24 & 0.21 & 0.15 \\
UNI & LP & 0.55 & 0.56 & 0.52 \\
UNI & FT\_last & 0.79 & 0.78 & 0.78 \\
\hline
\end{tabular}
}
\end{table}

Although experiments were conducted up to \texttt{flag\_limit} = 8192,
the overall scaling trends observed at this largest setting were consistent
with those at \texttt{flag\_limit} = 4096 for all evaluated architectures.
However, EfficientNet exhibited excessively long training and inference times
at \texttt{flag\_limit} = 8192 and was therefore terminated before completing
the full evaluation.
As a consequence, we report quantitative comparisons at
\texttt{flag\_limit} = 4096, where all task-specific and foundation models
could be evaluated under identical conditions.
Importantly, the relative performance ordering and the point at which
task-specific models surpassed foundation models were unchanged at
\texttt{flag\_limit} = 8192.

\subsection*{Data scale required to surpass foundation models}
To quantify how much labeled data are required for task-specific models to
match or exceed foundation models, we analyzed classification performance as a
function of \texttt{flag\_limit}
(Fig.~\ref{fig:scaling_flag_limit} and
Table~\ref{tab:classification_performance_4096}).
At small data scales (\texttt{flag\_limit} $\leq$ 512), all task-specific models
trained from scratch underperformed compared with foundation models using
linear probing (LP) or fine-tuning (FT\_last).
In this low-data regime, foundation models---particularly UNI with FT\_last---
exhibited a clear advantage, achieving macro-F1 scores above 0.8 while
task-specific models remained substantially lower.
As the training set size increased, performance gaps progressively narrowed.
For several convolutional architectures (ResNet, ConvNeXt, NIN), parity with
foundation models under LP was approached at intermediate scales
(\texttt{flag\_limit} $\approx$ 2048--4096), but these models did not consistently
exceed the performance of foundation models fine-tuned on the same task.
In contrast, CustomViT demonstrated a qualitatively different scaling behavior.
Already at \texttt{flag\_limit} = 2048, CustomViT matched or slightly exceeded
the performance of UNI under LP.
At \texttt{flag\_limit} = 4096, CustomViT clearly surpassed all LP-based
foundation models and approached the performance of UNI with FT\_last
(Table~\ref{tab:classification_performance_4096}).
Experiments conducted at \texttt{flag\_limit} = 8192 exhibited trends consistent
with those observed at \texttt{flag\_limit} = 4096
(Fig.~\ref{fig:scaling_flag_limit}).
However, EfficientNet required prohibitively long training and inference times
at this scale and was therefore terminated before completing a full evaluation.
For this reason, we report quantitative comparisons at
\texttt{flag\_limit} = 4096, which represents the largest data scale at which
all task-specific and foundation models could be evaluated under identical
conditions.
Notably, at \texttt{flag\_limit} = 8192, CustomViT continued to improve and
decisively exceeded the performance of UNI with FT\_last, confirming that the
observed crossover is robust to further data scaling.

\subsection*{Robustness to Blur}

\begin{figure}[htbp]
\centering
\includegraphics[width=\columnwidth]{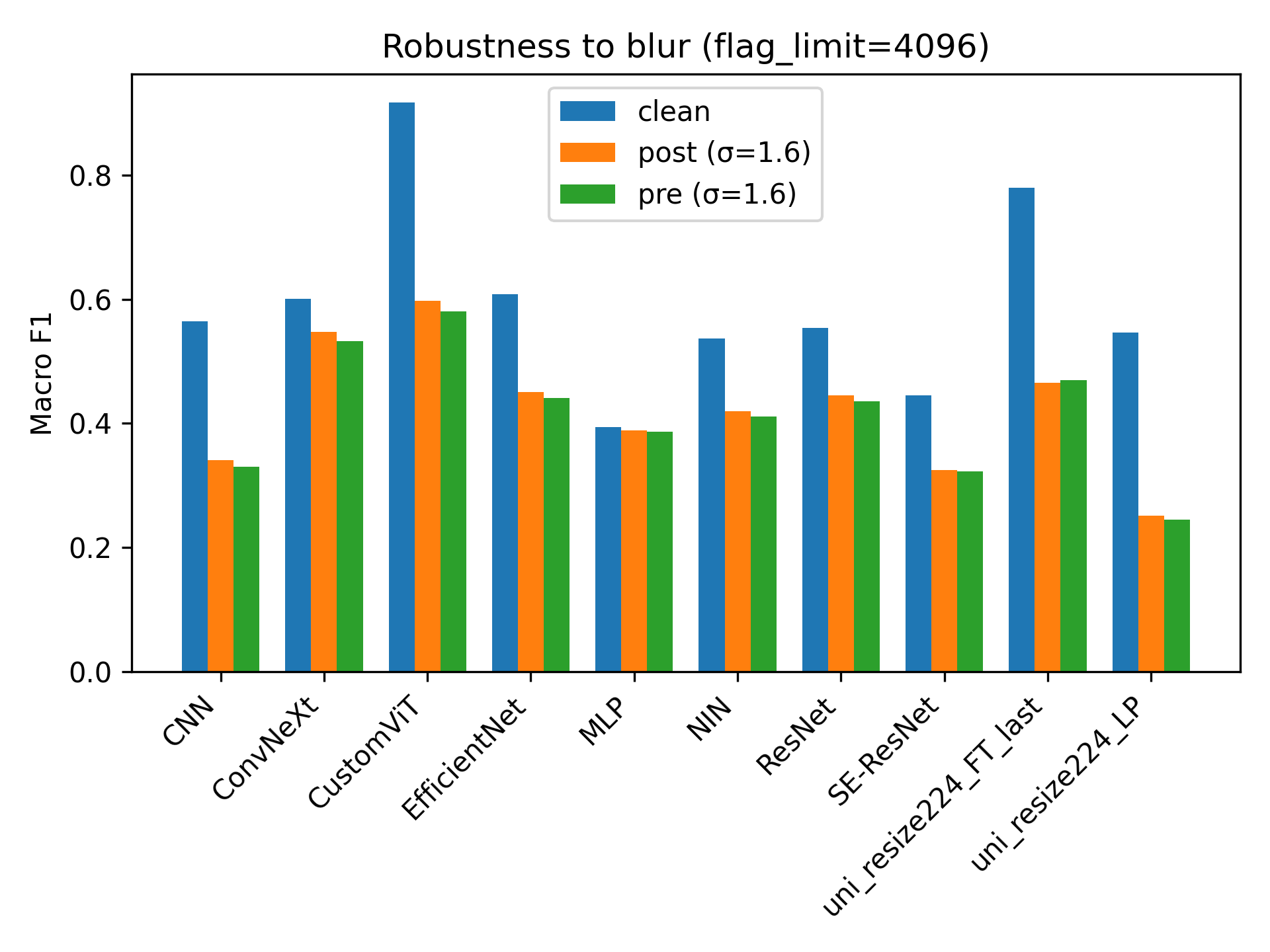}
\caption{
Robustness to strong blur ($\sigma = 1.6$) under a fixed annotation budget (flag\_limit = 4096).
Macro-F1 scores are shown for clean images (blue), post-resize blur (orange), and pre-resize blur (green).
Blur is applied either after resizing to $40 \times 40$ pixels (post) or before resizing with resolution-corrected strength (pre).
All models show pronounced performance degradation at this severity, with comparable sensitivity to pre- and post-resize blur.
}
\label{fig:robustness_4096_1.6}
\end{figure}

\begin{table}[htbp]
\centering
\caption{Robustness to blur at different severities ($\sigma$) under a fixed annotation budget (flag\_limit = 4096).
Macro-F1 scores and relative changes from the clean baseline ($\Delta$ Macro-F1) are reported.
For each blur type (pre-resize and post-resize), the label is shown only in the first row of the corresponding block.}
\label{tab:robustness_blur_4096}
\begin{tabular}{llrrrr}
\hline
Model & Blur & $\sigma$ & Macro-F1 & $\Delta$ Macro-F1 \\
\hline
ResNet & pre  & 0.1 & 0.553384 & -0.000187 \\
       &      & 0.2 & 0.553114 & -0.000457 \\
       &      & 0.4 & 0.548343 & -0.005228 \\
       &      & 0.8 & 0.526534 & -0.027037 \\
       &      & 1.6 & 0.435413 & -0.118158 \\
       & post & 0.1 & 0.553571 & 0 \\
       &      & 0.2 & 0.553465 & -0.000106 \\
       &      & 0.4 & 0.551806 & -0.001765 \\
       &      & 0.8 & 0.528185 & -0.025385 \\
       &      & 1.6 & 0.444646 & -0.108924 \\
\hline
CustomViT & pre  & 0.1 & 0.916980 & 0.000081 \\
          &      & 0.2 & 0.916004 & -0.000895 \\
          &      & 0.4 & 0.906016 & -0.010883 \\
          &      & 0.8 & 0.838967 & -0.077931 \\
          &      & 1.6 & 0.580491 & -0.336408 \\
          & post & 0.1 & 0.916899 & 0 \\
          &      & 0.2 & 0.916899 & 0 \\
          &      & 0.4 & 0.913003 & -0.003896 \\
          &      & 0.8 & 0.842737 & -0.074162 \\
          &      & 1.6 & 0.597622 & -0.319276 \\
\hline
UNI FT\_last & pre  & 0.1 & 0.780562 & 0.000626 \\
                   &      & 0.2 & 0.780303 & 0.000367 \\
                   &      & 0.4 & 0.765950 & -0.013987 \\
                   &      & 0.8 & 0.675980 & -0.103957 \\
                   &      & 1.6 & 0.469279 & -0.310658 \\
                   & post & 0.1 & 0.779937 & 0 \\
                   &      & 0.2 & 0.779956 & 0.000020 \\
                   &      & 0.4 & 0.779205 & -0.000731 \\
                   &      & 0.8 & 0.673225 & -0.106712 \\
                   &      & 1.6 & 0.464888 & -0.315048 \\
\hline
\end{tabular}
\end{table}

\begin{figure}[htbp]
\centering
\includegraphics[width=\columnwidth]{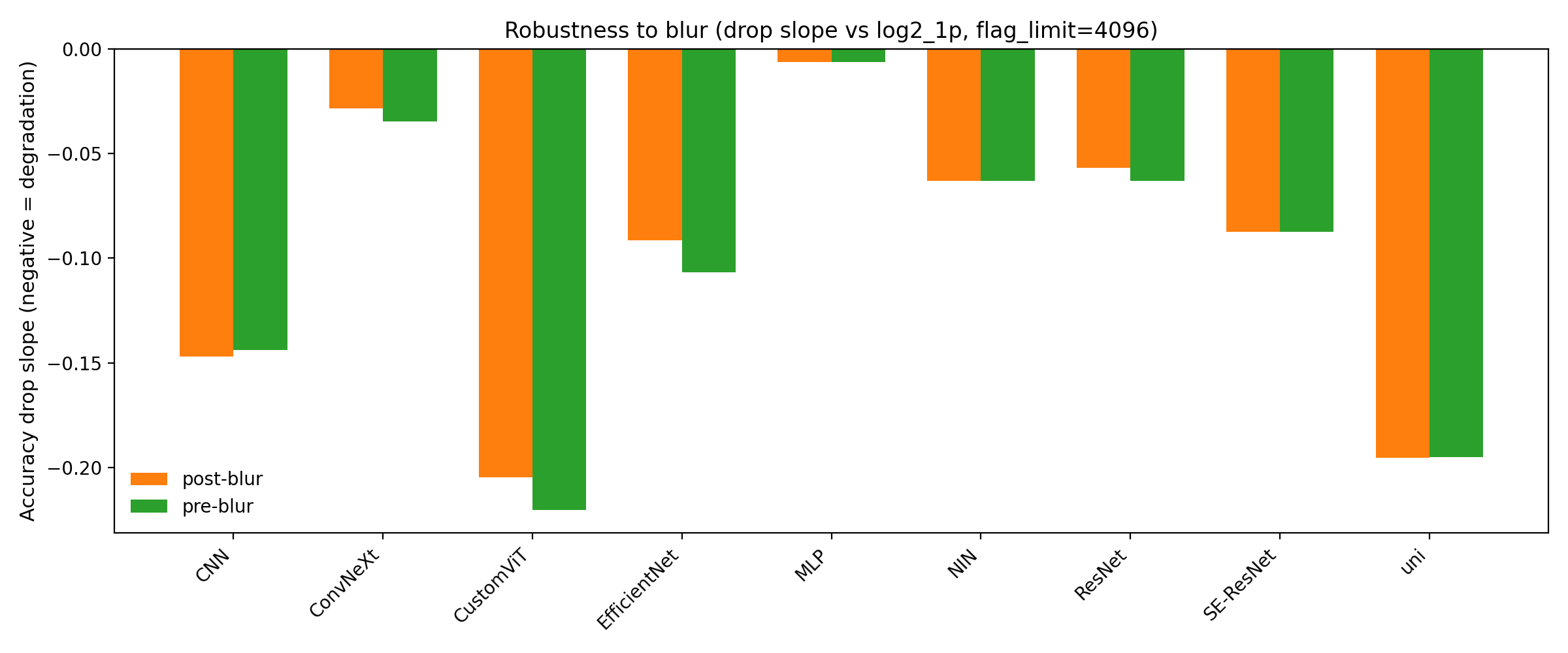}
\caption{
Accuracy degradation rate with increasing blur severity.
Bars indicate the slope of accuracy change per unit increase in $\log_2(1+\sigma)$,
with negative values representing faster performance degradation.
Pre-resize (optical) and post-resize (digital) blur are shown separately.
}
\label{fig:robustness_accuracy_drop_4096}
\end{figure}

We evaluated model robustness to image blur using both pre-resize (optical) and post-resize (digital) perturbations under a fixed annotation budget (flag\_limit = 4096).
Figure~\ref{fig:robustness_4096_1.6} summarizes classification performance under a strong blur condition ($\sigma = 1.6$), while Figure~\ref{fig:robustness_accuracy_drop_4096} quantifies the overall sensitivity to blur severity by estimating the performance drop slope across the full range of perturbations.
Detailed quantitative results across multiple blur levels are reported in Table~\ref{tab:robustness_blur_4096}.

As shown in Figure~\ref{fig:robustness_4096_1.6}, all models exhibit substantial degradation at high blur levels, regardless of whether blur is applied before or after resizing.
This trend is consistently observed across architectures and blur severities (Table~\ref{tab:robustness_blur_4096}).
The performance gap between pre-resize and post-resize blur is generally modest, indicating that most architectures are similarly affected by physically plausible optical blur and synthetic digital blur when the effective spatial scale is matched.
Notably, performance remains relatively stable at lower blur levels ($\sigma \leq 0.4$), whereas a clear decline emerges for $\sigma \geq 0.8$, consistent with a threshold-like behavior in blur sensitivity.

The extent of degradation varies across architectures, but the relative ordering of models remains largely preserved.
Vision Transformer–based models and strong convolutional baselines achieve higher absolute performance under both clean and corrupted conditions; however, they do not exhibit qualitatively superior robustness compared to simpler CNN architectures when evaluated on small ($40 \times 40$) cell patches.
This pattern is evident in both the absolute performance values (Table~\ref{tab:robustness_blur_4096}) and the aggregated robustness metrics (Figure~\ref{fig:robustness_accuracy_drop_4096}).
These findings suggest that higher-capacity or foundation-style models primarily benefit from improved clean accuracy rather than intrinsically enhanced resistance to blur.

To further characterize robustness, Figure~\ref{fig:robustness_accuracy_drop_4096} reports the drop slope of accuracy as a function of blur severity, estimated by linear regression on a log$_2(1+\sigma)$ scale.
This representation captures the approximately exponential decay in performance observed as blur strength doubles.
Across models, pre-resize and post-resize slopes are closely matched, reinforcing the conclusion that robustness is governed more by architectural inductive bias than by the specific stage at which blur is introduced.

Overall, these results indicate that blur robustness does not substantially differentiate foundation models from compact, task-specific architectures in the small-patch regime.
As training data volume increases, the reliance on large pre-trained models becomes less critical, and architectures explicitly designed for low-resolution inputs may offer a more appropriate trade-off between performance, robustness, and computational efficiency.

%% file: 06_discussion.tex
\subsection{Impact of Squeeze-and-Excitation Architecture on Small-Patch Learning}

To assess whether channel-wise attention mechanisms improve performance in
small-patch medical image classification, we evaluated a Squeeze-and-Excitation
enhanced variant of ResNet-D4 (SE-ResNet-D4) against the standard architecture.
At a training scale of \texttt{flag\_limit} = 4096, SE-ResNet-D4 consistently
underperformed ResNet-D4, achieving 0.53 accuracy and 0.47 macro-F1 compared
with 0.61 and 0.57, respectively (Table~\ref{tab:classification_performance_4096}).

Training dynamics revealed that SE-ResNet-D4 exhibited slower convergence and a
larger train--validation gap, suggesting impaired generalization.
We hypothesize that in settings characterized by small input size
(40$\times$40) and subtle morphological cues, aggressive channel reweighting
may suppress informative low-level features rather than enhance discriminative
capacity.
These findings indicate that squeeze-and-excitation mechanisms are not
universally beneficial and may be counterproductive for small-patch medical
image classification; consequently, SE-ResNet architectures were excluded from
further comparisons.

\subsection{EfficientNet under Small Patch Constraints}

EfficientNet achieved slightly higher classification performance than ResNet-D4
at \texttt{flag\_limit} = 4096 (Table~\ref{tab:classification_performance_4096}).
However, this marginal improvement came at the cost of severe scaling
limitations at larger training set sizes.

At \texttt{flag\_limit} = 8192, EfficientNet training became strongly
memory-bound due to the extensive use of depthwise separable convolutions and
squeeze-and-excitation blocks, whose fine-grained operations lead to frequent memory accesses and limit effective kernel fusion.
This resulted in non-linear increases in wall-clock time without proportional
gains in GPU utilization, ultimately requiring early termination of training.

These results indicate that, despite its moderate accuracy advantage,
EfficientNet is unsuitable for benchmarking data efficiency against foundation
models in this setting: its training cost grew disproportionately with dataset
size, making the data scale required to surpass foundation models impractically
large.

\subsection{Vision Transformer Architectures for Small Patch Learning}

In contrast to convolutional and channel-attention-based architectures, the
Vision Transformer--based model (CustomViT) exhibited consistently superior
scaling behavior and ultimately surpassed all foundation models at large data
scales.

This success can be attributed to several architectural properties suited to
small patch-based classification.
First, Vision Transformers operate on explicit patch tokens, preserving spatial
locality without prematurely collapsing fine-grained information through
repeated pooling.
Second, self-attention enables global spatial interaction without relying on
channel-wise recalibration; unlike SE blocks, it directly models relationships
between spatial regions, which is advantageous when class-discriminative signals
arise from subtle spatial co-occurrence patterns.
Third, Vision Transformers avoid strong translation invariance biases inherent
to convolutional architectures, retaining flexibility in capturing task-specific
spatial dependencies that may convey diagnostic information in medical images.

Notably, these benefits became apparent only beyond a moderate data scale.
At low \texttt{flag\_limit} values, CustomViT did not outperform foundation
models, consistent with the data-hungry nature of transformers.
However, once sufficient labeled data were available
(\texttt{flag\_limit} $\geq$ 2048), performance improved monotonically,
ultimately surpassing all foundation models at the largest evaluated scale.

\subsection{Limitations of ConvNeXt for Small Patch Learning}

ConvNeXt incorporates design principles inspired by Vision Transformers,
including large convolutional kernels, LayerNorm, and GELU activations.
While effective for large-scale natural image classification, our results
indicate that ConvNeXt architectures are poorly suited for small patch-based
medical image learning.

We evaluated a modified ConvNeXt variant adapted for 40$\times$40 inputs,
reducing the depthwise kernel size from 7$\times$7 to 5$\times$5 and adjusting
the stem convolution and downsampling operations accordingly.
Despite these modifications, the adapted ConvNeXt failed to achieve meaningful
improvements over conventional convolutional baselines, suggesting a fundamental
incompatibility rather than mere kernel size mismatch.

This limitation likely stems from several factors: ConvNeXt rapidly accumulates
large effective receptive fields through repeated depthwise convolution and
downsampling, leading to premature spatial information loss; its architecture
emphasizes channel-wise feature mixing via pointwise MLP layers, prioritizing
inter-channel relationships over spatial reconfiguration; and global average
pooling further compresses spatial information, disproportionately affecting
performance at severely constrained input resolutions.
These results highlight the importance of selecting architectures whose
inductive biases align with both the spatial scale and task characteristics.

\subsection{Foundation Models versus Task-Specific Architectures}

Foundation models provide substantial advantages at small data scales, with UNI
achieving F1 scores above 0.8 at \texttt{flag\_limit} $\leq$ 512 through
transfer learning.
However, at \texttt{flag\_limit} = 4096, CustomViT trained from scratch
(F1 = 0.92) decisively surpassed fine-tuned UNI (F1 = 0.78).

This crossover reflects architectural misalignment: foundation models are
pretrained on images substantially larger than 40$\times$40 pixels and rely on
multi-scale spatial features unavailable in small patches.
CTransPath exhibited markedly lower performance (F1 = 0.01--0.15), suggesting
that Swin Transformer's hierarchical windowed attention is poorly suited to
inputs lacking spatial hierarchy.
In contrast, CustomViT's patch-token representation operates effectively at
the 40$\times$40 scale without requiring multi-scale context.

The data scale required for CustomViT to surpass foundation models
(\texttt{flag\_limit} $\approx$ 2048--4096) remains experimentally feasible,
highlighting the practical viability of task-specific architectures when
input characteristics diverge from foundation-model pretraining regimes.
Note that foundation model evaluation was constrained by GPU memory limitations,
requiring reduced batch sizes (13 for linear probing, 5 for fine-tuning versus
80 for task-specific models) due to larger input size (224$\times$224).

\subsection{Robustness to Blur}

Across all architectures, classification performance remains largely stable under mild blur ($\sigma \leq 0.4$) and degrades rapidly only at higher blur levels ($\sigma \geq 0.8$), indicating a threshold-like sensitivity rather than gradual deterioration. This threshold-like behavior is consistently observed for both pre-resize and post-resize blur, suggesting that moderate blur does not substantially disrupt discriminative cell-level cues in small ($40 \times 40$) patches.

When the effective blur strength is normalized to the same spatial scale, the difference between pre-resize (optical) and post-resize (digital) blur is generally modest. Both absolute performance under strong blur (Figure~\ref{fig:robustness_4096_1.6}) and accuracy drop slopes across blur severities (Figure~\ref{fig:robustness_accuracy_drop_4096}) show closely matched trends. This indicates that, once scale effects are controlled, robustness is governed primarily by architectural inductive bias rather than by the stage at which blur is introduced.

Vision Transformer--based models achieve the highest absolute accuracy under both clean and blurred conditions. However, they also exhibit the steepest performance drops as blur severity increases. This highlights an important distinction: strong clean performance does not necessarily imply intrinsic robustness. In the small-patch regime, ViTs appear particularly sensitive to the loss of fine-grained spatial information caused by severe blur.

Overall, robustness patterns are broadly similar across architectures, and higher-capacity or foundation-style models do not demonstrate qualitatively superior blur resistance compared with compact convolutional baselines. Importantly, robustness analysis here is not intended as a primary benchmark, but rather as supporting evidence that, as training data volume increases, the advantages of large pre-trained models lie mainly in improved clean accuracy. For low-resolution cell-level analysis, task-specific architectures trained for this regime may offer a more appropriate trade-off between accuracy, robustness, and computational efficiency.

%% file: 07_conclusion.tex
This study demonstrates that cell-level pathological image classification on 40×40 pixel patches is feasible with appropriately designed architectures. CustomViT, a Vision Transformer optimized for small patches, achieved superior performance (F1 = 0.92) compared with foundation models while providing order-of-magnitude computational advantages (1.78 ms vs 24.89 ms inference time). Foundation models demonstrated strong performance in low-data regimes but exhibited limited scaling beyond 2,048 samples per class, indicating fundamental transferability constraints under extreme spatial downsampling. Channel-attention mechanisms proved detrimental under small-patch constraints, with both SE-ResNet-D4 and EfficientNet-B0 underperforming baseline architectures.

For cell-level pathological image analysis with datasets exceeding 4,096 samples per class, task-specific Vision Transformer architectures trained from scratch provide superior classification accuracy, substantially lower inference costs, and comparable robustness to foundation models. These findings challenge the prevailing assumption that foundation models universally outperform task-specific architectures in medical imaging, highlighting the importance of matching architectural design to domain-specific constraints.

%% file: 09_acknowlegements.tex
\section*{Acknowledgements}

We would 
like to express our sincere gratitude to the following individuals for their valuable contributions to the precise image tagging: To the graduate students and technical assistants from Kobe University who assisted with the annotation tasks: Abe T, Adachi Y, Agawa K, Ando M, Fukuda S, Imai M, Ito R, Kagiyama H, Konaka R, Miyake T, Mukoyama T, Okazoe Y, Takahashi T, Ueda Y and Yasuda K. We also extend our appreciation to the registered annotators who were engaged by the AMAIC: Adachi K, Akima J, Aoki S, Ichikawa K, Kanto T, Kawase Y, Kimura M, Miura R, Sirasawa H, Sotani K and Yuki A. Their professional and diligent efforts greatly enhanced the quality of the dataset.
This study is supported by the Grants-in-Aid for Scientific Research from the Ministry of Education, Culture, Sports, Science and Technology of Japan (MEXT; 24K10381 to TN, 23K08171 to KY, and 21K09167 to MF).

%% file: 10_declaration_of_generative_AI.tex
\section*{Declaration of generative AI and AI-assisted technologies in the writing process}

During the preparation of this work the authors used Claude Opus 4.5 (Anthropic) 
and ChatGPT 5.2 (OpenAI) to check for errors in analysis software code and to 
improve the language and readability of the manuscript text. After using these 
tools, the authors reviewed and edited the content as needed and take full 
responsibility for the content of the publication.

%% file: 20_appendix.tex
\subsection*{Training Dynamics and Class-wise Error Analysis}

To verify that the performance trends reported in the main Results are not artifacts of unstable optimization or premature overfitting, we provide additional diagnostic analyses of training dynamics and class-wise error structures for representative models.
Specifically, we examine a conventional convolutional baseline (ResNet), a task-specific transformer architecture (CustomViT), and a fine-tuned foundation model (UNI with FT\_last).
Training and validation curves are used to assess convergence behavior, while normalized confusion matrices provide insight into class-dependent error patterns.
These analyses are intended to support the reliability of the reported scaling results rather than to introduce additional performance comparisons.

\begin{figure}[htbp]
\centering
\includegraphics[width=\columnwidth]{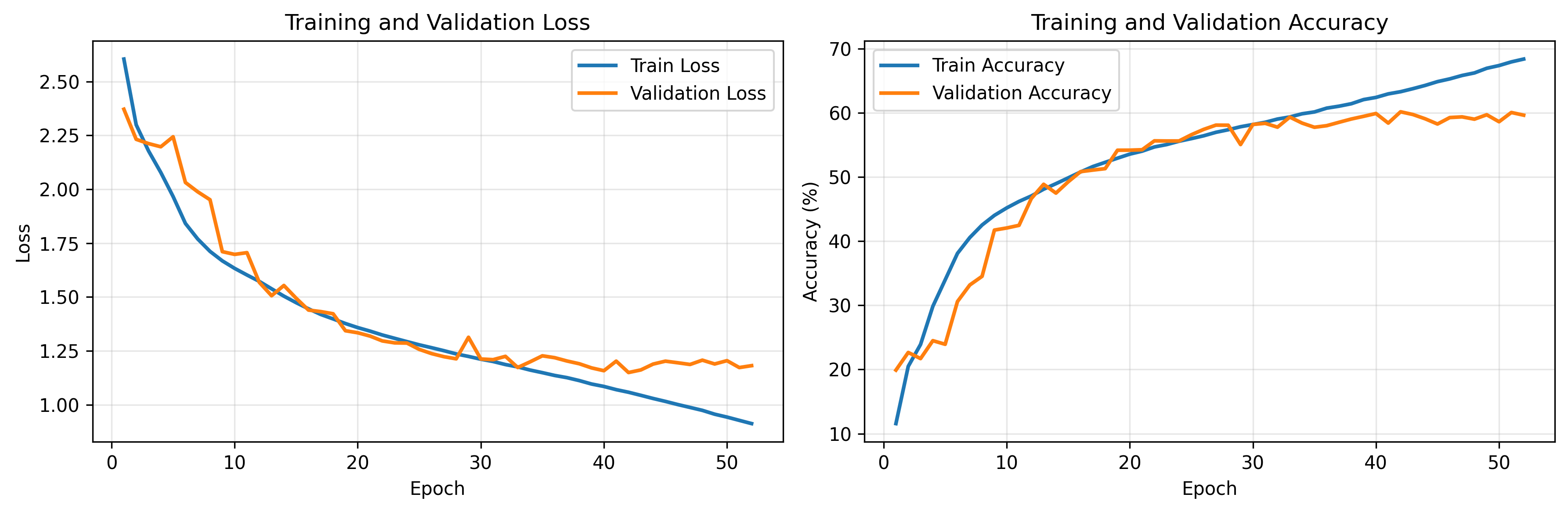}
\caption{
Training dynamics of ResNet over 50 epochs.
The left panel shows training and validation loss curves, exhibiting smooth exponential decay and early stabilization.
The right panel displays corresponding accuracy trajectories, with validation accuracy plateauing after approximately 30 epochs.
These trends indicate stable optimization and the absence of severe overfitting.
}
\label{fig:resnet_training}
\end{figure}

\begin{figure}[htbp]
\centering
\includegraphics[width=0.8\columnwidth]{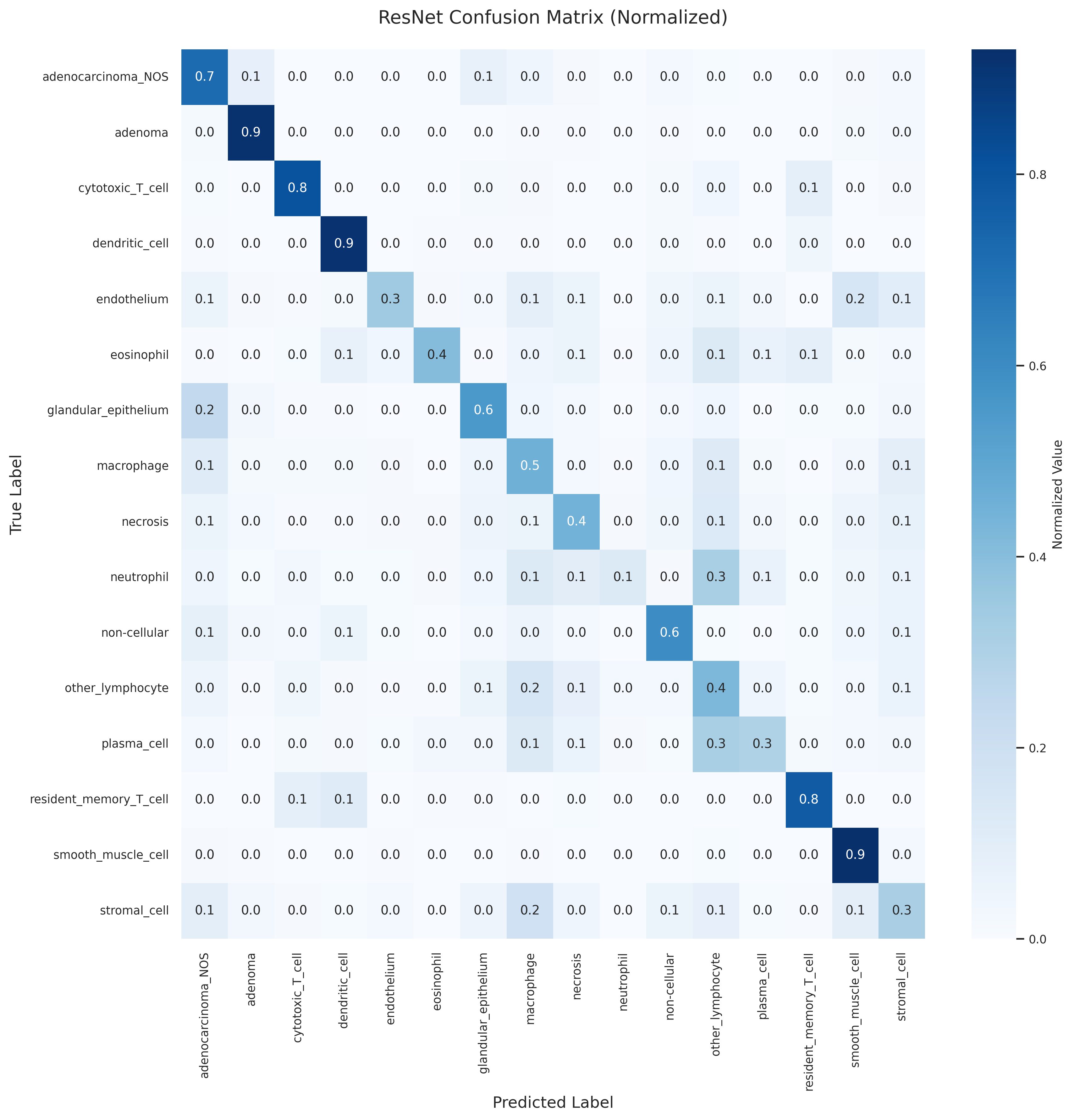}
\caption{
Normalized confusion matrix for ResNet classification across 16 cell types.
Diagonal entries represent correct classification rates, while off-diagonal entries indicate misclassification patterns.
Epithelial and structurally distinctive classes (e.g., adenocarcinoma NOS, adenoma, cytotoxic T cell, dendritic cell, smooth muscle cell) show strong diagonal dominance.
In contrast, morphologically heterogeneous or weakly discriminative classes (e.g., endothelium, neutrophil, plasma cell, stromal cell) exhibit increased confusion with related cell populations.
}
\label{fig:resnet_confusion}
\end{figure}

\paragraph{ResNet (Convolutional baseline)}
As a representative convolutional architecture, ResNet provides a baseline
for assessing optimization stability and class-wise error structure in the
small-patch regime.
As shown in Fig.~\ref{fig:resnet_training}, training and validation curves
exhibit smooth convergence without evidence of severe overfitting.
However, the confusion matrix (Fig.~\ref{fig:resnet_confusion}) reveals
substantial confusion among morphologically similar or heterogeneous classes,
highlighting the limitations of conventional convolutional features at this
resolution.

\begin{figure}[htbp]
\centering
\includegraphics[width=\columnwidth]{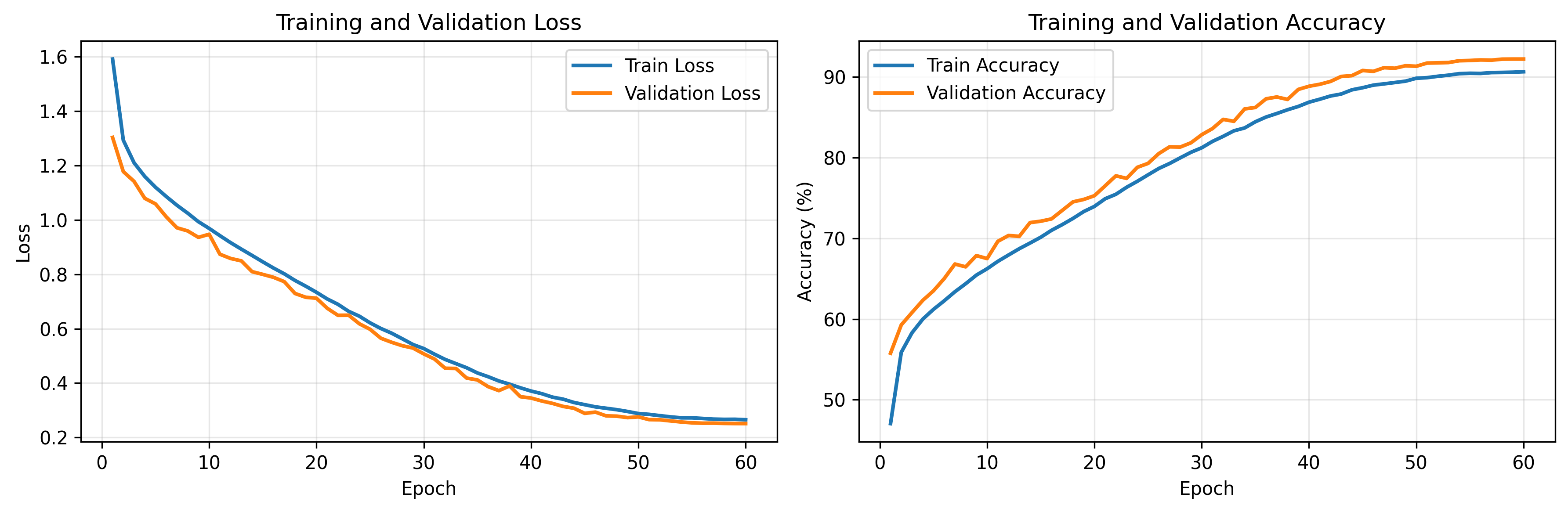}
\caption{
Training dynamics of CustomViT over 60 epochs.
Both training and validation loss decrease smoothly, accompanied by monotonic improvements in accuracy.
The close alignment between training and validation curves indicates stable convergence despite the higher model capacity.
Compared with convolutional baselines, CustomViT achieves substantially higher validation accuracy without evidence of optimization instability.
}
\label{fig:vit_training}
\end{figure}

\begin{figure}[htbp]
\centering
\includegraphics[width=0.8\columnwidth]{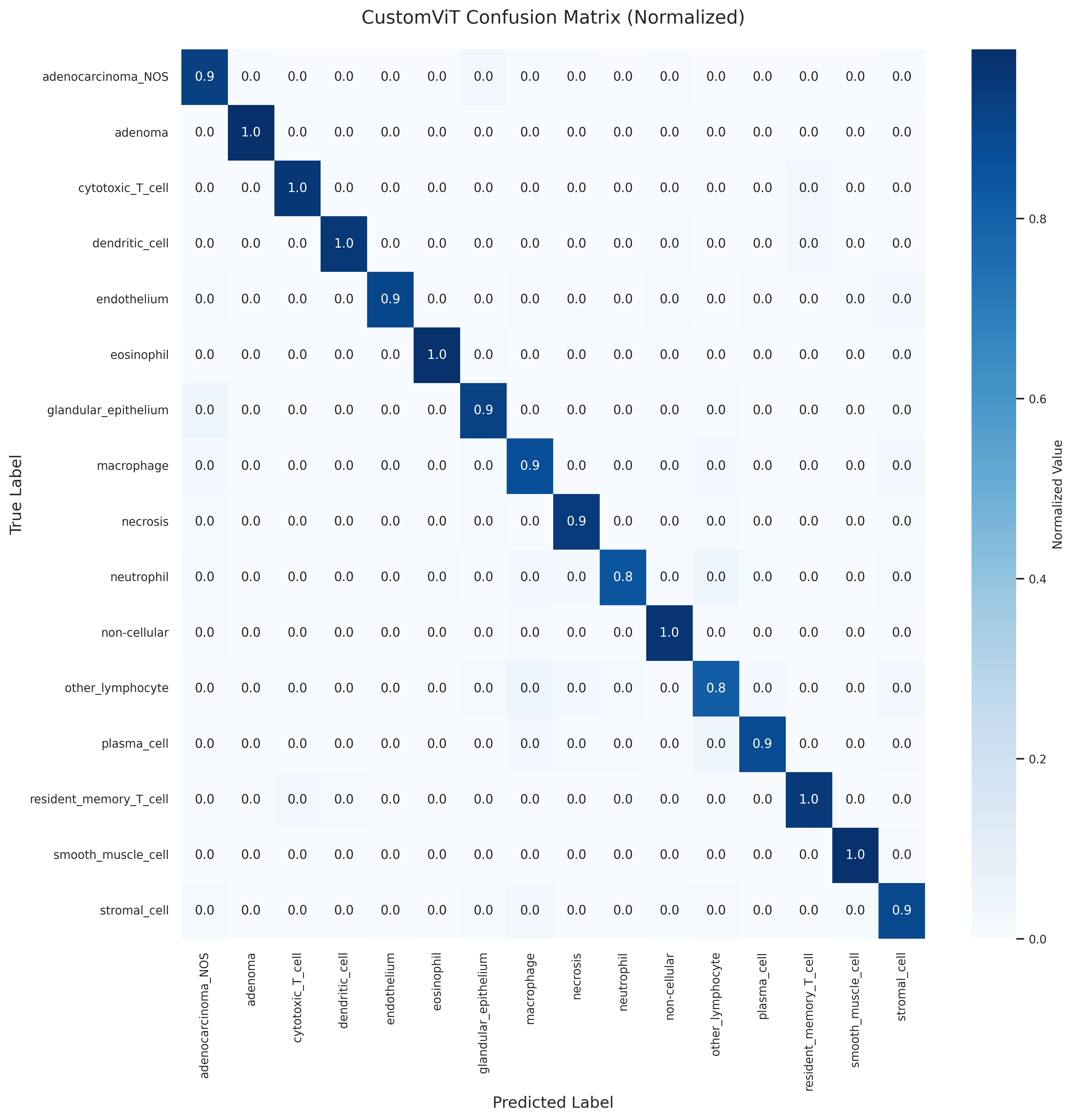}
\caption{
Normalized confusion matrix for CustomViT across 16 cell types.
Strong diagonal dominance is observed for nearly all classes, reflecting improved feature separability relative to convolutional architectures.
Residual confusion is primarily confined to biologically heterogeneous categories such as stromal cells and other lymphocytes, consistent with intrinsic label ambiguity rather than optimization failure.
}
\label{fig:vit_confusion}
\end{figure}

\paragraph{CustomViT (Task-specific transformer)}
CustomViT represents a task-specific transformer architecture optimized for
small-patch cell classification.
Figure~\ref{fig:vit_training} demonstrates stable convergence with monotonic
accuracy improvements.
The corresponding confusion matrix (Fig.~\ref{fig:vit_confusion}) shows strong
diagonal dominance across most cell types, indicating improved feature
separability relative to convolutional baselines.

\begin{figure}[htbp]
\centering
\includegraphics[width=\columnwidth]{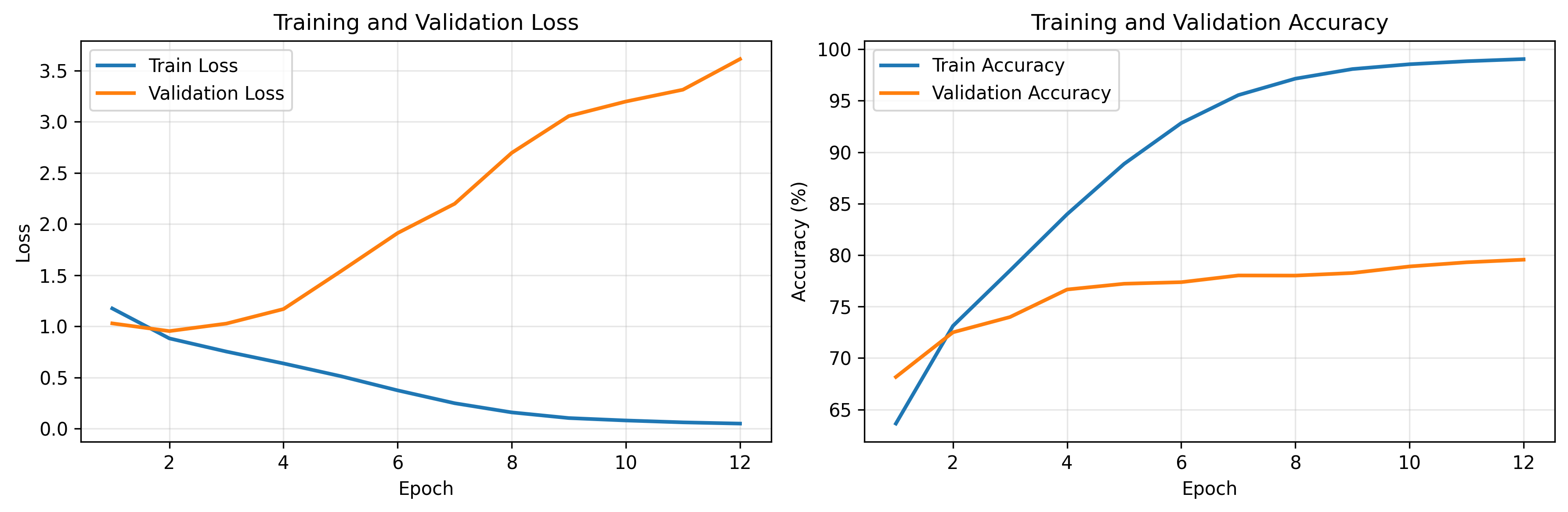}
\caption{
Training dynamics of UNI with fine-tuning of the last layers (FT\_last).
Rapid initial convergence is followed by stable refinement of both loss and accuracy.
The early saturation of training accuracy reflects the strong representational prior provided by large-scale self-supervised pretraining.
}
\label{fig:uni_training}
\end{figure}

\begin{figure}[htbp]
\centering
\includegraphics[width=0.8\columnwidth]{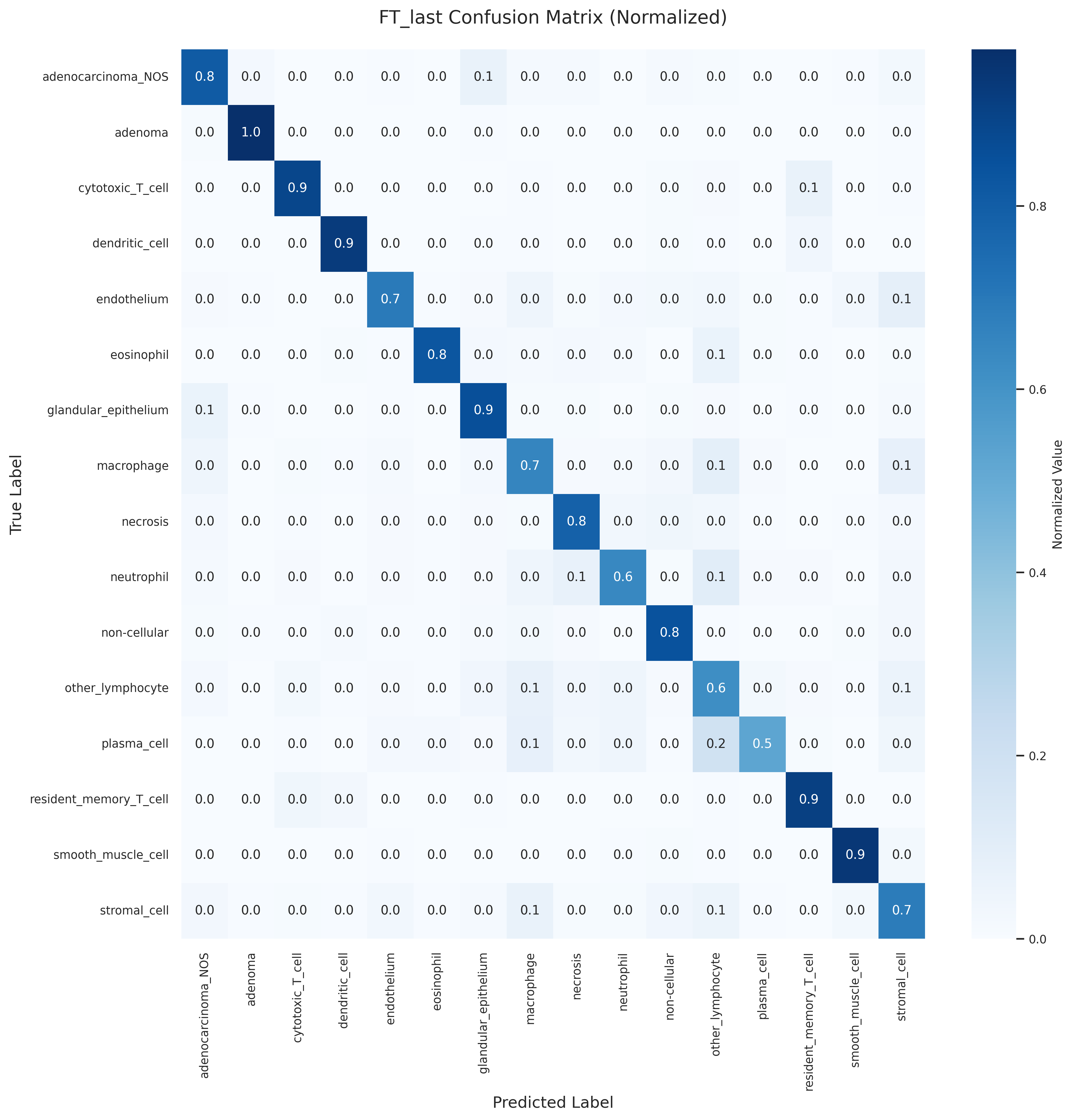}
\caption{
Normalized confusion matrix for UNI with FT\_last.
Overall classification accuracy is high, with well-separated epithelial and immune cell populations.
Remaining errors are concentrated in morphologically overlapping or weakly stained classes, suggesting that residual limitations arise from data characteristics rather than insufficient model capacity.
}
\label{fig:uni_confusion}
\end{figure}

\paragraph{UNI with FT\_last (Foundation model)}
UNI with fine-tuning of the last layers serves as a representative foundation
model with strong self-supervised pretraining.
As shown in Fig.~\ref{fig:uni_training}, optimization converges rapidly,
reflecting the influence of pretrained representations.
While validation accuracy continues to improve gradually, an increasing gap
between training and validation accuracy is observed.
This behavior is consistent with partial adaptation of a strong pretrained
representation to a fine-grained, small-patch classification task, resulting in
a form of task-mismatch--induced bias rather than classical high-variance
overfitting.
Consistent with this interpretation, the confusion matrix
(Fig.~\ref{fig:uni_confusion}) indicates that remaining classification errors are
concentrated in morphologically overlapping or weakly stained classes, suggesting
data- and label-driven limitations rather than optimization instability.

This behavior can be formally understood by considering the restriction
imposed on the effective hypothesis class under limited fine-tuning.

When transfer learning is performed via linear probing, where all pretrained
layers remain frozen, the effective hypothesis class is constrained to functions
of the form
\[
\mathcal{H}_{\text{fixed}}
=
\{ g_w \circ \phi_{\theta_0} \mid w \in W \},
\]
where $\phi_{\theta_0} : \mathcal{X} \rightarrow \mathbb{R}^d$ denotes a pretrained
feature extractor with fixed parameters, and
$g_w : \mathbb{R}^d \rightarrow \mathcal{Y}$ represents a trainable classification
head with parameters $w$.
Under this constraint, learning is limited to adjusting the decision function on
top of a fixed representation space.
This formulation applies directly to linear probing and, more generally, to
restricted fine-tuning schemes in which the feature extractor remains
approximately fixed.
Even if the pretrained feature map $\phi_{\theta_0}$ provides a rich and
high-dimensional representation, the target Bayes-optimal decision rule $f^\ast$
may not be realizable within $\mathcal{H}_{\text{fixed}}$ due to mismatch between
the pretraining task and the downstream classification objective.
As a consequence, an approximation gap may persist:
\[
\Delta_{\text{approx}} = R_{\mathcal{H}_{\text{fixed}}}^\ast - R^\ast \geq 0,
\]
where $R^\ast$ denotes the Bayes risk and
$R_{\mathcal{H}_{\text{fixed}}}^\ast$ is the minimum achievable risk within the
restricted hypothesis class.
When positive, this approximation error arises from the limited expressiveness of
the constrained hypothesis class and does not diminish with increasing sample
size, distinguishing it from estimation error associated with finite-sample
learning.

Taken together, these training dynamics and class-wise error analyses
support the interpretation that the observed performance differences and
scaling behaviors arise from systematic architectural characteristics and
data-related factors, rather than from optimization artifacts.